%% file: mm23.tex
\documentclass[sigconf]{acmart}
\usepackage[linesnumbered,ruled]{algorithm2e}
\usepackage{makecell}
\usepackage{arydshln}
\usepackage{booktabs}
\usepackage{color}

\AtBeginDocument{%
  }

\copyrightyear{2023}
\acmYear{2023}
\setcopyright{acmlicensed}
\acmConference[MM '23] {Proceedings of the 31st ACM International Conference on Multimedia}{October 29--November 3, 2023}{Ottawa, ON, Canada.}
\acmBooktitle{Proceedings of the 31st ACM International Conference on Multimedia (MM '23), October 29--November 3, 2023, Ottawa, ON, Canada}
\acmPrice{15.00}
\acmISBN{979-8-4007-0108-5/23/10}
\acmDOI{10.1145/3581783.3611735}

\begin{document}

\title{Improving Human-Object Interaction Detection \\ via Virtual Image Learning}
\renewcommand{\shorttitle}{Improving Human-Object Interaction Detection via Virtual Image Learning}

\author{Shuman Fang}
\affiliation{%
  \institution{Key Laboratory of Multimedia Trusted Perception and Efficient Computing, Ministry of Education of China, Xiamen University}
  \country{}
}
\email{fangshuman@stu.xmu.edu.cn}

\author{Shuai Liu}
\affiliation{%
  \institution{Key Laboratory of Multimedia Trusted Perception and Efficient Computing, Ministry of Education of China, Xiamen University}
  \country{}
}
\email{luckyliu@stu.xmu.edu.cn}

\author{Jie Li}
\affiliation{%
  \institution{Key Laboratory of Multimedia Trusted Perception and Efficient Computing, Ministry of Education of China, Xiamen University}
  \country{}
}
\email{lijie.32@outlook.com}

\author{Guannan Jiang}
\affiliation{%
  \institution{Intelligent Manufacturing Department, Contemporary Amperex Technology Co. Limited (CATL)}
  \country{}
}
\email{jianggn@catl.com}

\author{Xianming Lin}
\authornote{Corresponding author.}
\affiliation{%
  \institution{Key Laboratory of Multimedia Trusted Perception and Efficient Computing, Ministry of Education of China, Xiamen University}
  \country{}
}
\email{linxm@xmu.edu.cn}

\author{Rongrong Ji}
\affiliation{%
  \institution{Key Laboratory of Multimedia Trusted Perception and Efficient Computing, Ministry of Education of China, Xiamen University}
  \country{}
}
\email{rrji@xmu.edu.cn}


\input{sec0_abs.tex}



\maketitle

\input{sec1_intro.tex}
\input{sec2_rel_work.tex}
\input{sec3_method.tex}

\input{sec4_exp.tex}

\input{sec5_conclusion.tex}
\input{sec_acknowledgement.tex}

\bibliographystyle{ACM-Reference-Format}
\bibliography{bibtex}

\end{document}

%% file: sec0_abs.tex
\begin{abstract}
Human-Object Interaction (HOI) detection aims to understand the interactions between humans and objects,
which plays a curtail role in high-level semantic understanding tasks.
However, most works pursue designing better architectures to learn overall features more efficiently,
while ignoring the long-tail nature of interaction-object pair categories.
In this paper, 
we propose to alleviate the impact of such an unbalanced distribution via \textit{\textbf{V}irtual \textbf{I}mage \textbf{L}eaning} (VIL).
Firstly, a novel label-to-image approach, \textit{\textbf{Mu}ltiple \textbf{S}teps \textbf{I}mage \textbf{C}reation} (MUSIC),
is proposed to create a high-quality dataset that has a consistent distribution with real images.
In this stage, 
virtual images are generated based on prompts with specific characterizations and selected by multi-filtering processes.
Secondly,
we use both virtual and real images to train the model with the teacher-student framework.
Considering the initial labels of some virtual images are inaccurate and inadequate,
we devise an \textit{\textbf{A}daptive \textbf{M}atching-and-\textbf{F}iltering} (AMF) module to construct pseudo-labels.
Our method is independent of the internal structure of HOI detectors,
so it can be combined with off-the-shelf methods by training merely 10 additional epochs.
With the assistance of our method, multiple methods obtain significant improvements, 
and new state-of-the-art results are achieved on two benchmarks.

\end{abstract}

%% file: sec1_intro.tex
\section{Introduction}
\input{fig/overview.tex}
HOI detection is to comprehend the interactive relationships between humans and objects,
which can be denoted as a set of triplets $\langle$\textit{human, object, interaction}$\rangle$.
It has attracted the interest of many researchers due to its strong correlation with other vision tasks.
It not only contributes to other high-level semantic understanding tasks~\cite{tsai2015study_vqa,liu2020visual_vqa,bruni2016do_actreg,yan2022look_actreg} but also benefits from basic vision tasks~\cite{yang2017learning_pose,carion2020end,hu2021istr,hu2023you}.
However, the long-tail distributions for interactions and objects are common in the dataset. 
The combinatorial nature of HOIs further exacerbates the number gap between rare and non-rare categories~\cite{xu2019learning_longtail,hou2020visual_vcl,hou2021detecting_fcl,hou2021affordance_atl,wang2022chairs_odm}.
Models trained on such datasets only fit well in the common categories, while ignoring the rare ones.

To address the long-tail issue, re-sampling and re-weighting are designed to make detectors focus on tailed categories~\cite{hou2020visual_vcl,wang2022chairs_odm,zhang2021mining_cdn}.
But they underperform due to insufficient diverse features.
Some methods~\cite{xu2019learning_longtail,ji2020sgap,hou2021detecting_fcl} involve latent linguistic embeddings of rare categories to augment feature space,
which suffers from lacking visual representations yet. 
To get diverse visual features,
ATL~\cite{hou2021affordance_atl} expands training images by introducing extra off-the-shelf datasets.
Nevertheless, gathering large-scale relative images is challenging.
Generating virtual images becomes a straightforward idea.
Considering Stable Diffision (SD)~\cite{rombach2022high_sd} has succeeded in generating high-quality images,
we pursue a more suitable way for HOI detection to augment real datasets with it.
%

In this work, 
we propose a training framework termed \textit{\textbf{V}irtual \textbf{I}mage \textbf{L}eaning} (VIL).
The paradigm comparison of existing methods and ours are shown in Figure\,\ref{fig:framework}.
%
Firstly, to ensure the consistency of virtual and real images,
we propose a novel label-to-image generation approach,
named \textit{\textbf{Mu}ltiple \textbf{S}teps \textbf{I}mage \textbf{C}reation} (MUSIC).
%
Specially, we first create natural linguistic descriptions based on object and interaction categories.
Then, add characterizations for humans to improve the quality of virtual images
and constraints for scenes to ensure the authenticity of interactions.
The retouched textual descriptions are received by Stable Diffision~\cite{rombach2022high_sd} to acquire virtual images. 
To eliminate the uncertainty and instability in the generation process, 
we discard low-confidence virtual images with several filter mechanisms,
including scene similarity, instance existence, and human-object interactiveness.
With this, an unlimited number of virtual images with annotations can be obtained.

Considering the initial annotations of virtual images are noisy and incomplete,
we have to correct inaccurate bounding boxes and complement the missing HOI triplets.
Hence, we refer to the teacher-student framework
and propose \textit{\textbf{A}daptive \textbf{M}atching-and-\textbf{F}iltering} (AMF) module to create pseudo-labels for the virtual images.
To improve the accuracy of bounding boxes, 
the AMF module computes matching costs adaptively to avoid interference from noisy boxes.
And adaptive thresholds are used to pick up high-confidence predictions to recall HOI triplets in the images.
The student model learns from both virtual and real images,
supervised by the pseudo-labels and the ground-truth labels.
Meanwhile, the teacher model receives the student model's feedback
to improve the quality of pseudo-labels in the next iteration.
%

Our proposed method is simple, general, and orthogonal to existing methods.
By combining with our method, almost all off-the-shelf HOI detectors can be improved with only additional 10 training epochs.
%
To evaluate its efficiency,
we conduct extensive experiments with multiple off-the-shelf HOI detectors on two widely used datasets.
All experimented methods achieve relative gains and new state-of-the-art results are obtained on two datasets.
Ablation studies verify the contribution of each part of VIL.
Visualization experiments of virtual images and pseudo-labels explain the source of efficiency of our proposed MUSIC and AMF module.
Concretely, we summarize our contribution as follows:
\begin{itemize}
    \item We propose VIL, a model-agnostic framework, to boost the detection performance of existing methods.
    \item We devise a label-to-image generation approach named MUSIC, which can generate high-quality virtual images that have consistent distribution with real images.
    \item We design AMF, a pseudo-label generation module, to correct and supplement the initial labels of virtual images.
    \item By combining multiple methods with ours, the performance gains of all methods and the new state-of-the-art results indicate the efficacy of VIL.
\end{itemize}

%% file: fig/overview.tex
\begin{figure}[t]
\centering
\includegraphics[width=0.47\textwidth]{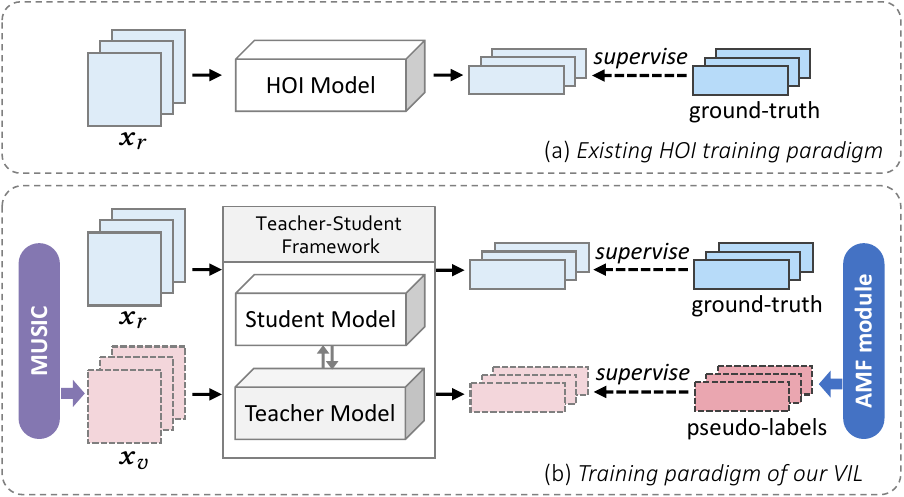}
\caption{Comparison for the HOI training paradigm.
(a) Most existing HOI detectors receive real images $\boldsymbol{x}_r$ to predict HOI triplets, 
supervised by ground truth of $\boldsymbol{x}_r$.
(b) Our VIL is trained with both virtual images $\boldsymbol{x}_v$ and real images $\boldsymbol{x}_r$, with the supervision of pseudo-labels and ground truth.
During this, virtual images are generated from MUSIC, and pseudo-labels are constructed by the AMF module.
}
\label{fig:framework}
\vspace{-10pt}
\end{figure}

%% file: sec2_rel_work.tex
\section{Related Work}

\subsection*{Category Bias Solution in HOI Detection}
The performance of many existing HOI methods~\cite{gao2018ican,ulutan2020vsgnet,liao2020ppdm,zhou2022human_distr,liao2022gen_genvlkt} is limited by the long-tail issue.
Methods proposed to address it can be categorized into three streams: re-sampling, re-weighting, and data space extension.

VCL~\cite{hou2020visual_vcl} emphasizes the long-tail issue for the first time.
Given an input image, VCL randomly samples another one and permutes interaction-object pair in these two images to obtain new combinations,
which somehow alleviates the long-tail dilemma.
%
ODM~\cite{wang2022chairs_odm} proposes another sampling strategy by alternately performing write-in and read-out stages.
The write-in stage dynamically updates the memory with rare categories' features, 
whereas the read-out part seeks to sample the far-distance features.

%
A dynamic re-weighting mechanism is proposed by CDN~\cite{zhang2021mining_cdn},
which amplifies the weight of rare categories during extra training epochs with a relatively small learning rate.
However, both re-sampling and re-weighting suffer from overfitting existing features.

To relieve overfitting to the existing rare representations, 
an effective solution is to expand the available space of data or features.
FCL~\cite{hou2021detecting_fcl}, based on word embeddings of interaction-object pair categories, 
generates virtual features from Gaussian noise to enrich feature space.
Instead of using latent features which lack visual representations,
ATL~\cite{hou2021affordance_atl} directly extends the original dataset with additional datasets~\cite{lin2014microsoft,shao2019objects365}.
Learning from such rich and varied knowledge, the model can get rid of unbalanced distribution.
%
However, such additional datasets are hard to obtain and still suffer from the limited number of images.
Consequently, what we pursue is to generate large-scale virtual images with distributions that are consistent with the original ones.

\subsection*{Data Augmentation Based on Stable Diffusion}
The development of the diffusion model~\cite{ho2020denoising_ddpm} has enabled current study, Stable Diffusion (SD)~\cite{rombach2022high_sd}, to produce high-quality images with remarkable progress.
It allow general conditioning inputs (\emph{e.g.}, text) to synthesize images.
%
%
%
Considering its success in image generation, numerous researchers have investigated its application in data augmentation~\cite{zhou2022towards_sd_app,ge2022dall_sd_app,he2022synthetic_sd_augment,azizi2023synthetic_sd_augment,sariyildiz2023fake_sd_augment,trabucco2023effective_sd_augment}.
Almost existing SD-based augmentation works focus on the instance-level. 
They commit to improving portraying realistic objects. 
However, images generated by such approaches are unsuitable for HOI detection. In addition to photo-realistic humans and objects, the synthetic images utilized for HOI identiìcation should also represent speciìc scene information and plausible interaction behaviors.
To this end, we propose MUSIC to address the inadequacies based on them.

%% file: sec3_method.tex
\input{fig/sd_generate.tex}
\section{Virtual Image Learning}

\subsection{Overview}
Our proposed VIL first generates a virtual dataset $\mathcal{D}_v$ based on the long-tail distribution,
where a label-to-image generation approach, MUSIC (described in Section\,\ref{sec:sd_gen}), is introduced.
Then, the teacher-student framework is adopted to train on both virtual dataset $\mathcal{D}_v$ and real dataset $\mathcal{D}_r$.
During this stage, we design an AMF module (presented in Section\,\ref{sec:sd_train}) to generate pseudo-labels for virtual images.
The student model learns knowledge supervised with the pseudo-labels of virtual images and the ground truth of real images.
Meanwhile, the teacher model is updated by the student model to provide more reliable pseudo-labels in the next iteration.

\subsection{Virtual Image Generation} \label{sec:sd_gen}
Due to the excellent image generation ability of Stable Diffusion~\cite{rombach2022high_sd}, recent researchers have an appetite for expanding datasets with it~\cite{zhou2022towards_sd_app,ge2022dall_sd_app}.
This text-to-image model can output the images corresponding to the given text prompt.
In this work, we also adopt it to generate virtual images based on HOI categories.
In particular,
given an interaction category $c_a$ and an object category $c_o$,
we first expand the pair $(c_a, c_o)$ into a description sentence: ``a photo of a person [$c_a$-ing] a/an [$c_o$]''.
Then, Stable Diffusion, denoted as $F_g$, receives the sentence as the prompt to generate the corresponding image.
We name this approach \textit{\textbf{D}irect \textbf{I}mage \textbf{C}reation} (DIC).
However, the virtual images created by DIC can not achieve satisfactory quality.
As Figure\,\ref{fig:sd_vis} shown,
we analyze the failure cases come from the following reasons:
1) for humans, only parts of human-body are visible;
2) for scenes, the background can not provide indicative information;
3) for interactiveness, there may not be interactions between humans and objects.
Based on these,
the MUSIC is proposed to handle the issues.
It consists of text refinement, Scene Similarity, Instance Existence, and Human-Object Interactiveness.
The overall procedure of MUSIC is depicted in Figure\,\ref{fig:sd_generate}.

\subsubsection{Text Refinement} \label{sec:sd_gen_text}
As mentioned above, 
the virtual images generated by plain descriptions only contain human body parts, lacking of human pose features.
We argue that human body parts alone can not reflect human appearance characteristics.
To close to the input prompt, 
model $F_g$ should generate an image with an integrated person by portraying his face or clothing.
Consequently, we prepare a word set $W_h$. The words in it are representative of human characteristics, like age, gender, occupation, \emph{etc}. 
MUSIC refines the label-based sentence by replacing ``person'' with a specific human characteristic word $w_h$ randomly sampled from $W_h$.
To make the scene indicative,
MUSIC also adds scene description into the prompt sentence.
In particular, another word set $W_s$ is also needed, which is composed of scene categories in Places365~\cite{zhou2017places}.
For each $(c_a, c_o)$, we count all possible scene categories, represented as $W_s^{(c_a, c_o)} \subseteq W_s$.
And the scene description $w_s$ is sampled from $W_s^{(c_a, c_o)}$.
To sum up,
the plain sentence is polished by adding human and scene depiction,
which can be unified as ``a photo of a/an [$w_h$] [$c_a$-ing] a/an [$c_o$] in the [$w_s$]''.
We denote the refined sentence as $t$ and send it into $F_g$ to generate the virtual image $\boldsymbol{x}_v$.
Next, multiple filtering stages will evaluate for $\boldsymbol{x}_v$ to determine whether to keep it or drop it.

\subsubsection{Scene Similarity} \label{sec:sd_gen_scene}
We argue that the scene in virtual image $\boldsymbol{x}_v$ should be similar to that in real images,
so we compute the cosine similarity of scene features for evaluation.
%
In this stage,
a scene prediction model $F_\text{scn}$~\cite{zhou2017places} is introduced to extra scene features.
The scene quality score $s_\text{scn}$ of $\boldsymbol{x}_v$ can be expressed as the following formula:
\begin{align}\label{equ:scene_sim}
    s_{\text{scn}} = 
    \mathop{\max}\limits_{\boldsymbol{x}_{r} \in \mathcal{D}_r}
    \frac{
        \big( F_\text{scn} \left( \boldsymbol{x}_v \right) \big)^{\top}
        \big( F_\text{scn} \left( \boldsymbol{x}_r \right) \big)
    }{
        \left\| F_\text{scn} \left( \boldsymbol{x}_v \right) \right\|_2 \cdot
        \left\| F_\text{scn} \left( \boldsymbol{x}_r \right) \right\|_2
    },
\end{align}
where $\mathcal{D}_{r}$ is the real dataset, and $\left\| \cdot \right\|_2$ is $\ell_2$ norm.
The virtual image $\boldsymbol{x}_v$ can be determined as passable to the scene similarity filtering 
only if $s_{\text{scn}}$ is greater than the threshold $\tau_\text{scn}$.
Otherwise, it will be abandoned.

\subsubsection{Instance Existence} \label{sec:sd_gen_inst}
To avoid target humans or objects being too over-occluded to be detected,
$\boldsymbol{x}_v$ needs to be verified by Instance Existence filtering.
Specifically, $\boldsymbol{x}_v$ is sent to a pre-trained object detector $F_\text{det}$~\cite{carion2020end} to obtain a set of predictions.
MUSIC selects all predictions predicted as ``person'' or $c_o$ with confidence greater than $\tau_\text{det}$.
Then split them into two candidate sets $\boldsymbol{B}_{h}$ and $\boldsymbol{B}_{o}$.
%
If the lengths of the two candidate sets are both greater than $0$,
it is considered that $\boldsymbol{x}_v$ clearly contains a human and the target object.
If not, such an image will be discarded.

\subsubsection{Human-Object Interactiveness} \label{sec:sd_gen_inter}
In this turn, 
we need to verify whether $\boldsymbol{x}_v$ accurately expresses the semantic information of the refined sentence $t$.
In particular,
MUSIC enumerates the bounding boxes in $\boldsymbol{B}_{h}$ and $\boldsymbol{B}_{o}$ to combine them into a set of human-object pairs.
Each pair can be denoted as $(\boldsymbol{b}_h , \boldsymbol{b}_o)$.
The bounding box can be further represented as $\boldsymbol{b}_{\xi} = \big[ x_1^{\xi}, y_1^{\xi}, x_2^{\xi}, y_2^{\xi} \big]^{\top}$, where $\xi \in \{ h,o \}$.
We mask the pixels in human or object regions to obtain:
\begin{equation}
\begin{gathered}
    \label{equ:mask_ho}
        \boldsymbol{x}_\text{mask}  = \mathbf{M} \odot \boldsymbol{x}_{v}, \\
        \mathbf{M}_{(i,j)} = 
            \left\{ 
                \begin{aligned}
                    &1, \text{if }
            i \in [x^{\xi}_1,x^{\xi}_2] \land j \in [y^{\xi}_1, y^{\xi}_2] \\
                    &0, \text{otherwise} \\
                \end{aligned}.
            \right.
\end{gathered}
\end{equation}
By traversing all pairs, we can get a masked image set $\boldsymbol{X}_\text{mask}$.

After that, 
MUSIC uses a visual-linguistic model $F_\text{clip}$~\cite{radford2021learning_clip} 
to compute the semantic similarity between each masked image and the refined sentence $t$.
The maximum similarity will be regarded as the score $s_\text{inter}$ of this filtering step:
\begin{align} \label{equ:clip}
    s_\text{inter} = \mathop{\max}\limits_{\boldsymbol{x}_\text{mask} \in \boldsymbol{X}_\text{mask}}
    F_\text{clip} \left( \boldsymbol{x}_\text{mask}, t \right).
\end{align}
The virtual image $\boldsymbol{x}_v$ can pass through this step if $s_\text{inter}$ satisfies the threshold $\tau_\text{inter}$.
And the human-object pair $(\hat{\boldsymbol{b}}_h, \hat{\boldsymbol{b}}_o)$ that corresponds to the maximum similarity 
is served as the annotation bounding boxes for $\boldsymbol{x}_v$.
Otherwise, $\boldsymbol{x}_v$ will be dropped.

The virtual dataset $\mathcal{D}_v$ consists of all virtual images passing through all three filtering processes, formulated by:
\begin{equation}
\begin{gathered}
    \mathcal{D}_v = \big\{
        \big( \boldsymbol{x}_{v}^i, {y}_{v}^i \big) 
    \big\}_{i=1}^{N_{v}}, \\ 
    {y}_v =
        \big( c_a, c_o, \hat{\boldsymbol{b}}_{o}, \hat{\boldsymbol{b}}_{h} \big)
        \in
        \left\{1,\cdots,C_a\right\} \times \left\{1,\cdots,C_o\right\} \times
        \mathbb{R}^{4} \times \mathbb{R}^4,
\end{gathered}   
\end{equation}
where $N_v$ is the size of the virtual dataset, $C_a$ and $C_o$ are the numbers of interaction categories and object categories in the real dataset $\mathbb{\mathcal{D}}_r$.
With the help of MUSIC, the distribution of the virtual dataset is consistent with that of the real dataset.
In Figure\,\ref{fig:sd_vis}, we show some examples generated by MUSIC.

\subsection{Virtual Image Training} \label{sec:sd_train}
We argue that the initial annotations of virtual images still show deficiencies.
On the one hand, 
some of them contain incorrect bounding boxes, which means there do not exist interactions between the located human and object.
On the other hand, 
the number of HOI triplets in initial annotations is insufficient.
Annotation in each image only includes one HOI triplet since we adopt a one-label-to-one-image strategy to generate virtual.
But in fact, there are multiple triplets for each image.
Hence, a pseudo-labels module should be designed to correct and complement the initial annotations
so that pseudo-labels can supervise the learning of the virtual images better.
To end this,
we propose the AMF module to generate high-quality pseudo-labels
and introduce the teacher-student framework to train the model by learning from both virtual images and real images.

\subsubsection{Pseudo-Labels Generation} \label{sec:mf_module}
As mentioned above,
our proposed AMF module aims to 1) correct the wrong bounding boxes and 2) supplement the missing HOI triplets.

To achieve the former objective,
we adaptively compute the matching cost to find the most similar prediction.
In particular,
given virtual image $\boldsymbol{x}_v$,
its corresponding initial annotation is $y_v=\big( c_a, c_o, \hat{\boldsymbol{b}}_{o}, \hat{\boldsymbol{b}}_{h} \big)$.
By sending $\boldsymbol{x}_v$ into the teacher model,
the predictions consist of the following four parts:
the human bounding boxes $\{ \tilde{\boldsymbol{b}}_h^i \mid \tilde{\boldsymbol{b}}_h^i \in \mathbb{R}^4 \}_{i=1}^{N}$,
the object bounding boxes $\{ \tilde{\boldsymbol{b}}_o^i \mid \tilde{\boldsymbol{b}}_o^i \in \mathbb{R}^4 \}_{i=1}^{N}$,
the probability of object classes $\{ \tilde{\boldsymbol{s}}_o^i \mid \tilde{\boldsymbol{s}}_o^i \in [0,1]^{C_o+1} \}_{i=1}^{N}$, 
and the probability of interaction classes $\{ \tilde{\boldsymbol{s}}_a^i \mid \tilde{\boldsymbol{s}}_a^i \in [0,1]^{C_a} \}_{i=1}^{N}$,
where $N$ is the size of the prediction set.
For the $i$-th prediction, we formulate the matching cost of classification ${H}_\text{cls}^i$ as the following:
\begin{equation}
\begin{aligned}
    &{H}_\text{cls}^i = {H}_a^i + {H}_o^i, \\
    &{H}_a^i = -\frac{1}{2} \Big(
        \tilde{\boldsymbol{s}}_{a}^i[c_a] + 
        \frac{1}{N-1} \sum_{k \in \{1 \cdots N\} \backslash \{c_a\} } 1-\tilde{\boldsymbol{s}}_{a}^i[k]
    \Big), \\
    &{H}_o^i = -\tilde{\boldsymbol{s}}_{o}^i[c_o],
\end{aligned}
\end{equation}
where $[\cdot]$ means index operation.
And the localization cost ${H}_\text{loc}^i$ can be formulated like the following:
\begin{equation}
\begin{aligned}
    &{H}_\text{loc}^i = {H}_\text{reg}^i + {H}_\text{iou}^i, \\
    &{H}_\text{reg}^i = \max \big\{
        \big\| \tilde{\boldsymbol{b}}_{h}^i - \hat{\boldsymbol{b}}_h \big\|_1,
        \big\| \tilde{\boldsymbol{b}}_{o}^i - \hat{\boldsymbol{b}}_o \big\|_1
    \big\}, \\
    &{H}_\text{iou}^i = \max \big\{
        1 - {GIoU}(\tilde{\boldsymbol{b}}_{h}^i, \hat{\boldsymbol{b}}_h),
        1 - {GIoU}(\tilde{\boldsymbol{b}}_{o}^i, \hat{\boldsymbol{b}}_o)
    \big\},
\end{aligned}
\end{equation}
where ${GIoU}(\cdot)$ is the generalized IoU~\cite{rezatofighi2019generalized}.
Considering the bounding boxes may be inaccurate while the classes are absolutely correct, 
we compute the overall matching cost by adaptively dropping the localization part, that is:
\begin{align}
    {H}^i = \left\{
        \begin{aligned}
            &{H}_\text{cls}^i, 
                \text{if } \min_{k} {H}_\text{cls}^k + {H}_\text{loc}^k > 0 \\
            &{H}_\text{cls}^i + {H}_\text{loc}^i, 
                \text{otherwise}
        \end{aligned}
    \right..
\end{align}
With this, we can adaptively find the nearest prediction by searching for ${\omega} = \arg\min {H}^i$ with the Hungarian algorithm~\cite{kuhn1955hungarian}.
To correct the initial bounding boxes, 
we replace $\hat{\boldsymbol{b}}_h$ with $\tilde{\boldsymbol{b}}_{h}^{\omega}$ and 
$\hat{\boldsymbol{b}}_o$ with $\tilde{\boldsymbol{b}}_{o}^{\omega}$, respectively.
Also, we set the $\tilde{\boldsymbol{s}}_a^{\omega}[c_a]$ as infinite
to guarantee this prediction can be picked up in the high-confidence filtering.

For the latter objective, 
\emph{i.e.}, the number of HOI triplets in the initial annotation is insufficient,
we select high-confidence predictions to supplement it.
Considering the confidence gap among different HOI detectors,
we suggest using an adapt threshold to select predictions.
Firstly, we estimate the average number of human-object pairs in each of virtual images and denote the number as $\kappa$.
When the teacher model is introduced, we calculate its prediction scores for all virtual images.
And we select $(\kappa \times N_v)$-th highest score as the threshold $\tau_\text{bin}$.
All predictions with interaction confidence higher than $\tau_\text{bin}$ are picked up, 
where pair-wise NMS~\cite{zhang2021mining_cdn} is introduced to remove duplicate predictions.
By defining a score binarization function $f_\text{bin} (\boldsymbol{s}) = \lceil \boldsymbol{s} - \tau_\text{bin} \rceil$,
we can get the final pseudo-labels:
\begin{align}
    \tilde{{y}}_v = 
    \big\{
        \big( 
            f_\text{bin}(\tilde{\boldsymbol{s}}_{a}^{i}), 
            c_o, 
            \tilde{\boldsymbol{b}}_{h}^{i},
            \tilde{\boldsymbol{b}}_{o}^{i}
        \big)
        \mid
        \max \tilde{\boldsymbol{s}}_{a}^{i} > \tau_\text{bin}
    \big\}.
\end{align}
With high-quality pseudo-labels, the student model can learn from virtual images much more effectively.

\subsubsection{Teacher-Student Framework}
Inspired by Omni-DETR~\cite{wang2022omni},
we apply strong augmentation $T^{s}$ and weak augmentation $T^{w}$ to both virtual images $\boldsymbol{x}_v$ and real images $\boldsymbol{x}_r$.
And for virtual images $\boldsymbol{x}_v$,
we additionally design a random padding augmentation to avoid the model overfitting the large-area bounding boxes,
which can be formulated as:
\begin{align}
    T^{p}(\boldsymbol{x}_v, y_v) = \left\{
        \begin{aligned}
            &\text{RandomPad}(\boldsymbol{x}_v, y_v), \text{if }
                S_{\hat{\boldsymbol{b}}_{h}} > \frac{1}{2}S_{\boldsymbol{x}_v},p \leq 0.5 \\
            &(\boldsymbol{x}_v, y_v), \text{otherwise}
        \end{aligned}
        \right.,
\end{align}
where $\text{RandomPad}(\cdot)$ represents the random padding function,
the variable $p \sim {U}(0,1)$ is introduced to control the transformation rate, 
and $S_{\hat{\boldsymbol{b}}_{h}}$ and $S_{\boldsymbol{x}_v}$ are areas of the human bounding box and the virtual image.
Thus,
for each virtual data $({\boldsymbol{x}_v}, y_v) \in \mathcal{D}_v$, 
we can get $({\boldsymbol{x}_{v}^{ps}}, y_{v}^{ps})$ and $({\boldsymbol{x}_{v}^{pw}}, y_{v}^{pw})$,
where the superscript $ps$ means apply $T^{p}$ and $T^{s}$ successively, and similarly for $pw$.
For each real image $({\boldsymbol{x}_r}, y_r) \in \mathcal{D}_r$, they are transformed into $({\boldsymbol{x}_{r}^{s}}, y_{r}^{s})$ and $({\boldsymbol{x}_{r}^{w}}, y_{r}^{w})$.

In the training stage,
$\boldsymbol{x}_{v}^{pw}$ will be sent into the teacher model $\mathcal{F}_{t} (\boldsymbol{x}, \theta_{t})$ to predict pseudo-labels $\tilde{y}_{v}^{pw}$, 
which is described in Section\,\ref{sec:mf_module}.
Then we restore $\tilde{y}_{v}^{pw}$ by the inverse transformation of $T^{w}$
and apply the strong augmentation the same as $\boldsymbol{x}_{v}^{ps}$
to get the transformed pseudo-labels $\tilde{y}_{v}^{ps}$.
Finally, the student model $\mathcal{F}_{s} (\boldsymbol{x}, \theta_{s})$ is trained with $\boldsymbol{x}_{v}^{ps}$, $\boldsymbol{x}_{r}^{s}$, and $\boldsymbol{x}_{r}^{w}$ 
under the supervision of $\tilde{y}_{v}^{ps}$, ${y}_{r}^{s}$, and ${y}_{r}^{w}$, respectively.
The total loss function to optimize it can be represented as:
\begin{equation}
\begin{aligned}
    {L}_\text{total} &= 
                    \sum_{\boldsymbol{x}_v \in \mathcal{D}_v} {L}_\text{hoi}
                        \Big(
                            \mathcal{F}_{s}(\boldsymbol{x}_v^{ps}), \tilde{y}_{v}^{ps}
                        \Big) \\
                    &+ 
                    \sum_{\boldsymbol{x}_r \in \mathcal{D}_r} 
                        {L}_\text{hoi} \Big(
                            \mathcal{F}_{s}(\boldsymbol{x}_r^{s}), {y}_{r}^{s}
                        \Big) + 
                        {L}_\text{hoi} \Big(
                            \mathcal{F}_{s}(\boldsymbol{x}_r^{w}), {y}_{r}^{w}
                        \Big),
\end{aligned}
\end{equation}
where $L_\text{hoi} (\cdot, \cdot)$ denotes the loss function utilized in the off-the-shelf HOI detectors.

Meanwhile,
the teacher model should be updated by the exponential moving average (EMA)~\cite{tarvainen2017mean_ema} from the student model:
\begin{align}
    \theta_{t} \gets \alpha \theta_{t} + (1-\alpha) \theta_{s},
\end{align}
where $\alpha$ is a hyperparameter empirically set to a number close to 1 to keep the robustness in the teacher model.

%% file: fig/sd_generate.tex
\begin{figure*}[t]
\centering
\includegraphics[width=0.85\textwidth]{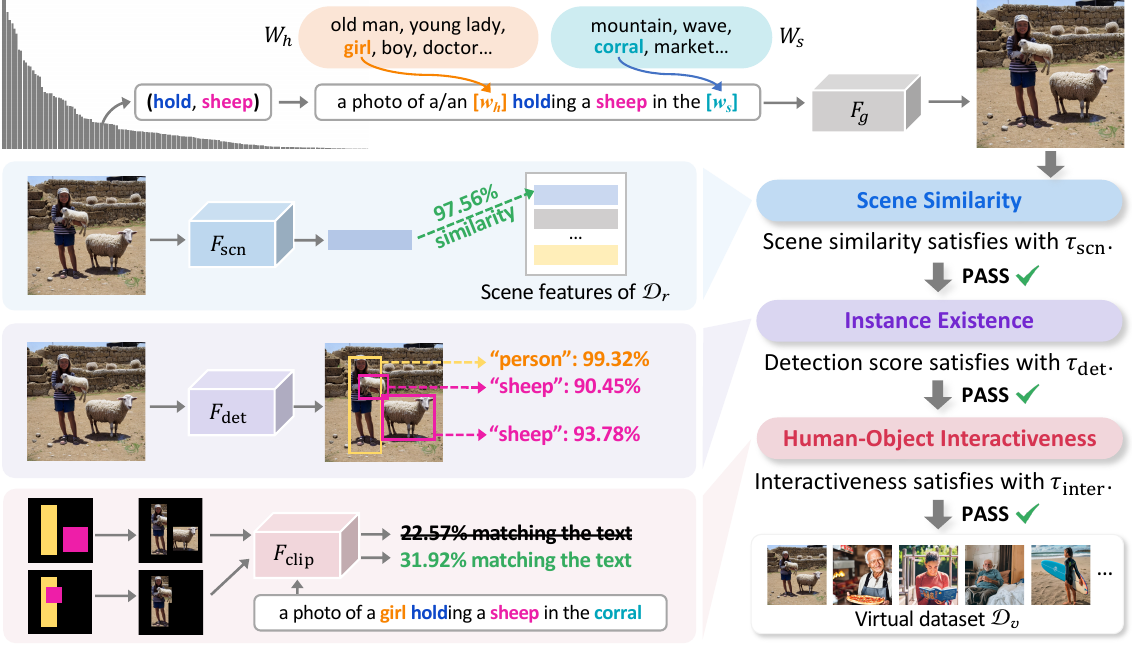}
\caption{
Overview of MUSIC.
Given a category $(c_a, c_o)$,
MUSIC firstly expands it into a plain linguistic sentence and refines it by adding human and scene depictions.
A text-to-image model $F_g$ receives the refined sentence as the prompt to generate a virtual image.
Then the image
will be evaluated by Scene Similarity, Instance Existence, and Human-Object Interactiveness
to determine whether to keep it or not.
}
\label{fig:sd_generate}
\end{figure*}

%% file: sec4_exp.tex
\section{Experiment}
\subsection{Experiment Setup}
\subsubsection{Dataset and Metric}
We follow previous works to evaluate performance on two public benchmarks, 
\emph{i.e.}, HICO-Det~\cite{chao2018learning_hico} and V-COCO~\cite{gupta2015visual_vcoco}.
We compute the mean average precision (mAP) to report experimental results.
A prediction is a true positive if the predicted human and object boxes have IoUs larger than 0.5 with the corresponding ground truth 
and the predicted HOI category is also correct.

In the HICO-Det dataset, 
there are 38,118 images for training and 9,658 for testing.
Images are annotated with 80 object classes and 117 action classes.
The HOI category is defined as an interaction-object pair (\emph{e.g.}, ``eat pizza'').
There are two settings for evaluation: {Default} and {Known Object}.
The Default setting requires evaluating all images, while Known Object only tests images containing the target object class.
In each setting, HICO-Det also provides three evaluation sets, \emph{i.e.}, Full, Rare, and Non-Rare, 
which are divided based on the frequency of categories.

V-COCO dataset contains 5,400 images in the trainval set and 4,946 in the test set.
Images are annotated with 80 object classes and 29 interaction classes. 
Four interaction classes (\emph{i.e.}, stand, walk, run, and smile) are neglected for evaluation 
since they are not associated with semantic roles. 
The HOI category is defined as an interaction class.
We report role mAP in two scenarios,
where scenario 1 needs to predict the cases in which humans interact with no objects 
while scenario 2 ignores these cases.

\input{tab/longtail.tex}
\input{tab/exp_res.tex}

\subsubsection{Implementation Details}
In the virtual image generation process,
we set the thresholds $\tau_\text{scn}$, $\tau_\text{det}$, and $\tau_\text{inter}$ as 0.9, 0.9, and 0.3, respectively.
For the HICO-Det dataset, 
the number of virtual images for each interaction-object pair category is set to 40 per rare category and 10 per non-rare category.
For the V-COCO dataset,
we first split all object-action pair categories into two groups:
the minority with less than 10 instances in the train set and the majority with 10 or more.
And we generate 30 and 15 virtual images for each minority and majority category, respectively.

During the training stage, $\alpha$ of EMA is set to 0.9996, the same as~\cite{tarvainen2017mean_ema,wang2022omni}.
Considering that our VIL is orthogonal with most existing methods, we conduct experiments by applying ours to them. 
Hence, we set the rest hyperparameters the same as the methods to be combined. 
We initialize the student and teacher model by loading parameters from the corresponding pre-trained method
and train the framework with 10 epochs.
The learning rate of the backbone and the other parts are set to the same as those after decay in the to-be-combined methods.
Considering virtual images are introduced due to category bias, 
we freeze the parameters except for classification heads when training with virtual images to prevent overfitting.

\subsection{Comparison with Long-tail Methods}

In Table~\,\ref{tab:longtail}, we compare our VIL with existing long-tail methods with QPIC~\cite{tamura2021qpic} as the baseline.
%
%
Note that ``+CDN'' means only applying the re-weighting technique in CDN.
From the table, only ATL brings the performance decline to QPIC. Since ATL is proposed based on traditional two-stage HOI methods, it needs the multi-stream structure to fuse features from additional datasets flexibly. Such a design makes it incompatible with transformer-based methods. Instead, our method is model-agnostic and suitable for almost all HOI methods.
Compared with the other methods, the improvement from our VIL also outstands those from others by a large margin on all sets.
We owe this to the various visual features provided by virtual images, 
which is the shortage of re-sampling~\cite{wang2022chairs_odm} and re-weighting~\cite{zhang2021mining_cdn} techniques.
Moreover,
by taking the SOTA method, GEN-VLKT~\cite{liao2022gen_genvlkt}, as baseline,
our method still outperforms existing works.
We also visualize the improvements from CDN and our VIL in Figure\,\ref{fig:improve}.
From the figure, the more rare the category is, the more improvement the baseline obtains.
And our improvement under rare categories is far more than CDN.
We conclude that our method can address long-tail problem much more effectively.

\input{fig/improve.tex}

\subsection{Improvement on Existing Works}

To evaluate the efficiency and generalization of our VIL, 
we select five representative methods to conduct combination experiments,
which are
one classic method QPIC~\cite{tamura2021qpic},
two relatively recent works that introduced extra knowledge OCN~\cite{yuan2022detecting_ocn} and DOQ~\cite{qu2022distillation_doq}, 
and two SOTA methods GEN-VLKT~\cite{liao2022gen_genvlkt} and DisTR~\cite{zhou2022human_distr}.

We first conduct experiments on the HICO-Det test set and report results in Table\,\ref{tab:hico}.
All methods achieve performance gains after combining with VIL.
For example,
QPIC with the help of VIL, improved by 1.47 mAP and 1.56 mAP under the Default and the Known Object setting,
surpassing many recent methods\cite{liu2022interactiveness_phrasehoi,wang2022chairs_odm,park2022consistency_cpc,iftekhar2022look_ssrt}.
%
OCN and DOQ get 1.08 and 0.85 mAP gains, respectively,
outperforming almost all existing works.
Similarly, the improvement of DisTR achieves 0.91 mAP.
The state-of-the-art method on HICO-Det, GEN-VLKT, can also be further enhanced to achieve a new SOTA result.
Note that in the Rare set under the two different settings, it acquires considerable margins of 1.33 and 2.13 mAP, with relative improvements achieving 4.55\% and 6.50\%.
We attribute the improvements on the Rare set to our long-tail-based design, 
which enables the model to pay more attention to features in rare categories.

For performance on the V-COCO test set, we still combine our VIL with the five works, which are QPIC, DOQ, OCN, GEN-VLKT, and DisTR.
The results are also reported in Table\,\ref{tab:hico}, where all methods are relatively improved.
Specifically, the early work, QPIC, attains 0.6 and 0.9 mAP in Scenario 1 and 2,
making it competitive with those latter works~\cite{liu2022interactiveness_phrasehoi,zhang2022efficient_upt}.
The other three recent works,
DOQ, OCN, and GEN-VLKT, 
are enhanced by 0.8, 0.7, and 0.7 mAP, respectively.
DisTR, the SOTA method on the V-COCO dataset, earns more significant improvement by 1.2 and 1.3 mAP in Scenario 1 and 2.
It achieves outstanding results of 67.3 and 69.7 mAP in the two scenarios,
refreshing the SOTA results.

\subsection{Ablation Study}
\input{tab/abla_music.tex}
\input{tab/abla_amf.tex}

Firstly, to verify the efficiency of each part of MUSIC,
we conduct ablation experiments for it and report the results in Table~\,\ref{tab:abla_music},
where the header ``Text'', ``Scene'', ``Instance'', and ``Interactiveness'' are corresponding the part depicted in Section~\,\ref{sec:sd_gen_text}, ~\,\ref{sec:sd_gen_scene}, ~\,\ref{sec:sd_gen_inst}, and ~\,\ref{sec:sd_gen_inter}.
As shown results, the performance degradations verify the necessity of each part. Also, the performance plummets dramatically when getting rid of Text Refinement (row 4). We argue that the characterizations of humans play a quite crucial role in virtual images, which is in line with intuition.
And the three last parts serve to validate the elements that need to be portrayed.
%

\input{fig/sd_vis.tex}
\input{fig/pseudo_labels.tex}

Then, we prove the effectiveness of the AMF module by removing ``Matching'' and ``Filtering'' one by one.
The results are shown in Table~\,\ref{tab:abla_amf}.
Without Filtering (row 1), learning one virtual image is supervised by only one HOI annotation. 
The insufficient annotations lead to a 0.2 mAP decline. 
By removing the Matching part (row 2), the further decline illustrates the detrimental impact of noisy bounding box annotations.
Moreover, the performance degradation in the two rows also illustrates that the AMF module can construct more reliable pseudo-labels to guide the student model.

\subsection{Qualitative Results} \label{sec:vis}
\subsubsection{Visualization for Virtual Images} 

The ablation study has demonstrated that the virtual images generated by DIC may impair the performance of existing methods,
whereas our MUSIC does the opposite.
To demonstrate this more intuitively,
we visualize the virtual images generated by these two approaches for comparison in Figure~\,\ref{fig:sd_vis},
where images at the first row are results from DIC and the rest are from MUSIC.
It is clear that the DIC images primarily lack human bodies, meaningful sceneries, or interactive human-object pairs.
For example,
all shown images only contain human hands. Among those, the hand in column 1 is severely occluded, and that in column 2 is illegible.
And images in columns 2, 5, and 8 miss the background information, while such information has been proved to be crucial to the understanding of interaction~\cite{wang2019deep_contextual,tamura2021qpic,zhang2021spatially_scg,qu2022distillation_doq}.
Additionally, for the image in column 6, the interaction ``lay'' between the human and the couch is not distinguished enough.
In contrast, the images generated via MUSIC can successfully address the aforementioned issues.
All images can perfectly convey the semantic information of the specified categories,
which benefits from the multiple filtering stages to guarantee the quality of virtual images.
Moreover,
the images have diversity in the appearance characteristics of the humans, the posture of the human bodies, and the scene where the interaction occurs.
In particular,
MUSIC can depict two different cases, ``cut hair with scissors'' and ``cut paper with scissors'', based on the category ``(cut\_instr, scissors)''.
For the category ``(lay, couch)'', various lying postures are constructed.
Apart from these, for the categories ``(read, book)'', ``(talk, cell\_phone)'', and ``(hit\_instr, tennis\_racket)'', 
MUSIC can provide a variety of reasonable scenes.
We owe these to the introduced textual descriptions that direct model $F_g$ to create various virtual images.

\subsubsection{Visualization for Pseudo-Labels}
Considering the noisy initial annotations created during the generation stage, 
AMF module is proposed to construct more reliable pseudo-labels.
Many initial annotations are with erroneous bounding boxes or lack sufficient HOI triplets.
We visualize some examples in Figure~\,\ref{fig:pseudo_labels}.
In the first row,
the initial annotation localizes the ``person'' incorrectly. 
It is the boy that looking at a pizza, not the man in the background.
In this case, we expect the pseudo-label to correct the human bounding box.
At the beginning epochs, 
the pseudo-labels successfully fix the bounding box of the human but conflate a bunch of pizzas.
As the training progresses, with the human bounding box keeping precise,
the bounding box of the pizza instance gradually moves closer to the correct direction, and finally locates exactly.
In the second row,
the initial annotation ignores the interaction between ``person'' and``baseball''.
As the pseudo-labels show, 
the interactive relationship is dug out at the 1st epoch.
But this relationship is lost again due to the instability of predictions.
As the teacher model is continuously updated,
this initially missed interactive triplet can be pointed out stably after the 6th epoch.
We emphasize the necessity of pseudo-labels,
which can provide better and richer supervised information for the student model.

%% file: tab/longtail.tex
\begin{table}[t]
\centering
\caption{Comparison with existing long-tail methods by combining QPIC~\cite{tamura2021qpic} on the HICO-Det dataset.
The content in parentheses indicates the performance improvements.
}

\label{tab:longtail}

\resizebox{0.475\textwidth}{!}{
\begin{tabular}{
    p{95pt}ccc
}
\toprule
\textbf{Method}  &
Full &
Rare &
Non-rare \\ \midrule






QPIC &
29.07 & 21.85 & 31.23 \\

QPIC + ATL~\cite{hou2021affordance_atl} & 
28.87 ($-$0.20) & 21.67 ($-$0.23) & 31.03 ($-$0.20) \\

QPIC + ODM~\cite{wang2022chairs_odm} & 
29.26 ($+$0.19) & 22.07 ($+$0.22) & 31.41 ($+$0.18) \\

QPIC + CDN~\cite{zhang2021mining_cdn} & 
29.40 ($+$0.33) & 21.96 ($+$0.11) & 31.63 ($+$0.40) \\


QPIC + VIL (ours) &
\textbf{30.54 ($+$1.47)} & \textbf{23.34 ($+$1.49)} & \textbf{32.69 ($+$1.46)} \\
\hline\hline 
GEN-VLKT &
33.75 & 29.25 & 35.10 \\


GEN-VLKT + ODM~\cite{wang2022chairs_odm} & 
33.82 ($+$0.07) & 29.59 ($+$0.34) & 35.08 ($-$0.02) \\

GEN-VLKT + CDN~\cite{zhang2021mining_cdn} & 
33.87 ($+$0.12) & 29.35 ($+$0.10) & 35.22 ($+$0.12) \\


GEN-VLKT + VIL (ours) &
\textbf{34.21 ($+$0.46)} & \textbf{30.58 ($+$1.33)} & \textbf{35.30 ($+$0.20)} \\
\bottomrule
\end{tabular}
}
\end{table}

%% file: tab/exp_res.tex
\begin{table*}[t]
\centering
\caption{Performance improvement on both HICO-Det and V-COCO datasets.
Each letter in the Feature column stands for \textbf{A}: Appearance/Visual feature, \textbf{S}: Spatial features, \textbf{L}: Linguistic feature of label semantic embeddings, and \textbf{P}: Human pose feature.
* signifies results reproduced with the official implementation codes.
The performance improvements in the Full and Rare sets are marked with \textbf{RED} and \textbf{BLUE}, respectively.
}

\label{tab:hico}

\resizebox{1.0\textwidth}{!}{
\begin{tabular}{
    l p{65pt}<{\centering} p{45pt}<{\centering}
    ccc ccc cc
}

\toprule

\multicolumn{3}{l}{} & 
\multicolumn{6}{c} {\textbf{HICO-Det}} &
\multicolumn{2}{c} {\textbf{V-COCO}} \\

\multicolumn{1}{l}{} & \multicolumn{1}{l}{} & \multicolumn{1}{l}{} &
\multicolumn{3}{c}{Default} &
\multicolumn{3}{c}{Known Object} &
\multicolumn{2}{c}{} \\ \cmidrule(r){4-6} \cmidrule(r){7-9} \cmidrule(r){10-11} 

\textbf{Method} & Backbone & Feature &
Full & Rare & Non-rare &
Full & Rare & Non-rare &
Scenario 1 & Scenario 2 \\ \midrule

DRG~\cite{gao2020drg_drg}
    & ResNet-50-FPN
    & A+S+P+L
    & 19.26 & 17.74 & 19.71 & 23.40 & 21.75 & 23.89 
    & 51.0  & - \\
VSGNet~\cite{ulutan2020vsgnet}
    & ResNet-152
    & A+S
    & 19.80 & 16.05 & 20.91 & -     & -     & -     
    & 51.8  & - \\
PPDM~\cite{liao2020ppdm}
    & Hourglass-104
    & A
    & 21.73 & 13.78 & 24.10 & 24.58 & 16.65 & 26.84 
    & - & - \\
GG-Net~\cite{zhong2021glance_ggnet}
    & Hourglass-104
    & A
    & 23.47 & 16.48 & 25.60 & 27.36 & 20.23 & 29.48 
    & - & - \\ 
SCG~\cite{zhang2021spatially_scg}
    & ResNet-50-FPN
    & A+S
    & 31.33 & 24.72 & 33.31 & 34.37 & 27.18 & 36.52 
    & 54.2  & 60.9  \\ [1pt]
%
PhraseHOI~\cite{liu2022interactiveness_phrasehoi}
    & ResNet-50
    & A+L
    & 29.29 & 22.03 & 31.46 & 31.97 & 23.99 & 34.36 
    & 57.4  & - \\
CPC~\cite{park2022consistency_cpc}
    & ResNet-50
    & A
    & 29.63 & 23.14 & 31.57 & - & - & - 
    & 63.1  & 65.4  \\
SSRT~\cite{iftekhar2022look_ssrt}
    & ResNet-50
    & A+L
    & 30.36 & 25.42 & 31.83 & - & - & - 
    & 63.7  & 65.9  \\
HQM~\cite{zhong2022towards_hqm}
    & ResNet-50
    & A
    & 31.34 & 26.54 & 32.78 & - & - & - 
    & 63.6  & - \\
UPT~\cite{zhang2022efficient_upt}
    & ResNet-50
    & A+S
    & 31.66 & 25.94 & 33.36 & 35.05 & 29.27 & 36.77 
    & 59.0  & 64.5  \\
CDN~\cite{zhang2021mining_cdn}
    & ResNet-50
    & A
    & 31.78 & 27.55 & 33.05 & 34.53 & 29.73 & 35.96 
    & 61.7  & 63.8  \\
SDT~\cite{wang2022distance_sdt}
    & ResNet-50
    & A
    & 32.45 & 28.09 & 33.75 & 35.95 & 31.30 & 37.34 
    & 60.3  & 65.7  \\
\hline\hline 
QPIC~\cite{tamura2021qpic}
    & ResNet-50
    & A
    & 29.07 & 21.85 & 31.23 & 31.68 & 24.14 & 33.93 
    & 58.8  & 61.0  \\
\quad+\textit{VIL (ours)} 
    & ResNet-50
    & A
    & 30.54 \textcolor{red}{($+$1.47)} & 23.34 & 32.69 
    & 33.24 \textcolor{red}{($+$1.56)} & 25.13 & 35.66 
    & 59.4 \textcolor{red}{($+$0.6)} & 61.9 \textcolor{red}{($+$0.9)} \\
\hdashline[3pt/3pt]\rule{-3pt}{10pt}
OCN~\cite{yuan2022detecting_ocn}
    & ResNet-50
    & A+L
    & 30.91 & 25.56 & 32.51 & 33.68* & 28.27* & 35.30* 
    & 64.2  & 66.3  \\
\quad+\textit{VIL (ours)} 
    & ResNet-50
    & A+L
    & 31.99 \textcolor{red}{($+$1.08)} & 26.67 & 33.58 
    & 34.75 \textcolor{red}{($+$1.07)} & 29.49 & 36.32
    & 64.9 \textcolor{red}{($+$0.7)} & 67.0 \textcolor{red}{($+$0.7)} \\
\hdashline[3pt/3pt]\rule{-3pt}{10pt}
DOQ~\cite{qu2022distillation_doq}
    & ResNet-50
    & A+L
    & 31.55 & 26.75 & 32.99 & 34.11* & 29.25* & 35.55* 
    & 63.5  & 65.9* \\
\quad+\textit{VIL (ours)} 
    & ResNet-50
    & A+L
    & 32.40 \textcolor{red}{($+$0.85)} & 27.95 & 33.73 
    & 34.99 \textcolor{red}{($+$0.88)} & 30.19 & 36.42
    & 64.3 \textcolor{red}{($+$0.8)} & 67.1 \textcolor{red}{($+$1.2)} \\
\hdashline[3pt/3pt]\rule{-3pt}{10pt}
DisTR~\cite{zhou2022human_distr}
    & ResNet-50
    & A
    & 31.93* & 27.26* & 33.32* & 34.62* & 29.53* & 36.14* 
    & 66.4*  & 68.6*  \\
\quad+\textit{VIL (ours)} 
    & ResNet-50
    & A
    & 32.84 \textcolor{red}{($+$0.91)} & 28.04 & 34.27 
    & 35.63 \textcolor{red}{($+$1.01)} & 30.53 & 37.15
    & \textbf{67.6} \textcolor{red}{($+$1.2)} & \textbf{69.9} \textcolor{red}{($+$1.3)} \\
\hdashline[3pt/3pt]\rule{-3pt}{10pt}
GEN-VLKT~\cite{liao2022gen_genvlkt}
    & ResNet-50
    & A+L
    & 33.75 & 29.25 & 35.10 & 36.78 & 32.75 & 37.99 
    & 64.6* & 66.8* \\
\quad+\textit{VIL (ours)} 
    & ResNet-50
    & A+L
    & \textbf{34.21} \textcolor{red}{($+$0.46)} & \textbf{30.58} & \textbf{35.30}
    & \textbf{37.67} \textcolor{red}{($+$0.89)} & \textbf{34.88} & \textbf{38.50}
    & 65.3 \textcolor{red}{($+$0.7)} & 67.7 \textcolor{red}{($+$0.9)} \\

\bottomrule
\end{tabular}
}

\end{table*}

%% file: fig/improve.tex
\begin{figure*}[t]
\centering
\includegraphics[width=1.0\textwidth]{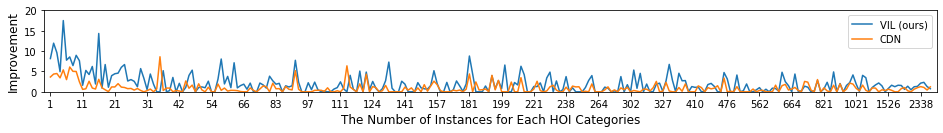}
\caption{
Comparison improvements with our VIL and CDN~\cite{zhang2021mining_cdn} 
by taking QPIC~\cite{tamura2021qpic} as baseline method on the HICO-Det dataset under the Default setting. 
We sort the categories by their frequency. The horizontal axis is the number of instances in the train set, and the vertical axis is the mean improvements of categories with the same number of instances.
}
\label{fig:improve}
\end{figure*}

%% file: tab/abla_music.tex
\begin{table}[t]
\centering

\caption{Ablation experiments for MUSIC approach (depicted in Section~\,\ref{sec:sd_gen}) on the V-COCO test set.
}

\label{tab:abla_music}

\resizebox{0.49\textwidth}{!}{
\begin{tabular}{
    c 
    cccc
    cc
}
\toprule
\#Row & Text & Scene & Instance & Interactiveness & Scenario 1 & Scenario 2  \\ \midrule
\textbf{Ours} 
    & \checkmark & \checkmark & \checkmark & \checkmark
    & \textbf{59.4} & \textbf{61.9} \\ 
1   & \checkmark & \checkmark & \checkmark & 
    & 59.2 ($-$0.2) & 61.6 ($-$0.3) \\ 
2   & \checkmark & \checkmark &            & 
    & 59.8 ($-$0.5) & 61.3 ($-$0.6) \\
3   & \checkmark &            &            & 
    & 58.7 ($-$0.7) & 61.1 ($-$0.8) \\
4   &            &            &            & 
    & 58.2 ($-$1.2) & 60.7 ($-$1.2) \\
\bottomrule
\end{tabular}
}
\vspace{-1em}
\end{table}

%% file: tab/abla_amf.tex
\begin{table}[t]
\centering

\caption{Ablation experiments for AMF module (depicted in Section~\,\ref{sec:sd_train}) on V-COCO test set.
}

\label{tab:abla_amf}

\resizebox{0.37\textwidth}{!}{
\begin{tabular}{
    c 
    cc
    cc
}
\toprule
\#Row & Matching & Filtering & Scenario 1 & Scenario 2  \\ \midrule
\textbf{Ours} 
    & \checkmark & \checkmark
    & \textbf{59.4} & \textbf{61.9} \\ 
1   & \checkmark &
    & 59.2 ($-$0.2) & 61.6 ($-$0.3) \\ 
2   &            & 
    & 58.8 ($-$0.6) & 61.4 ($-$0.4) \\
\bottomrule
\end{tabular}
}
\vspace{-1em}
\end{table}

%% file: fig/sd_vis.tex
\begin{figure*}[t]
\centering
\includegraphics[width=1.\textwidth]{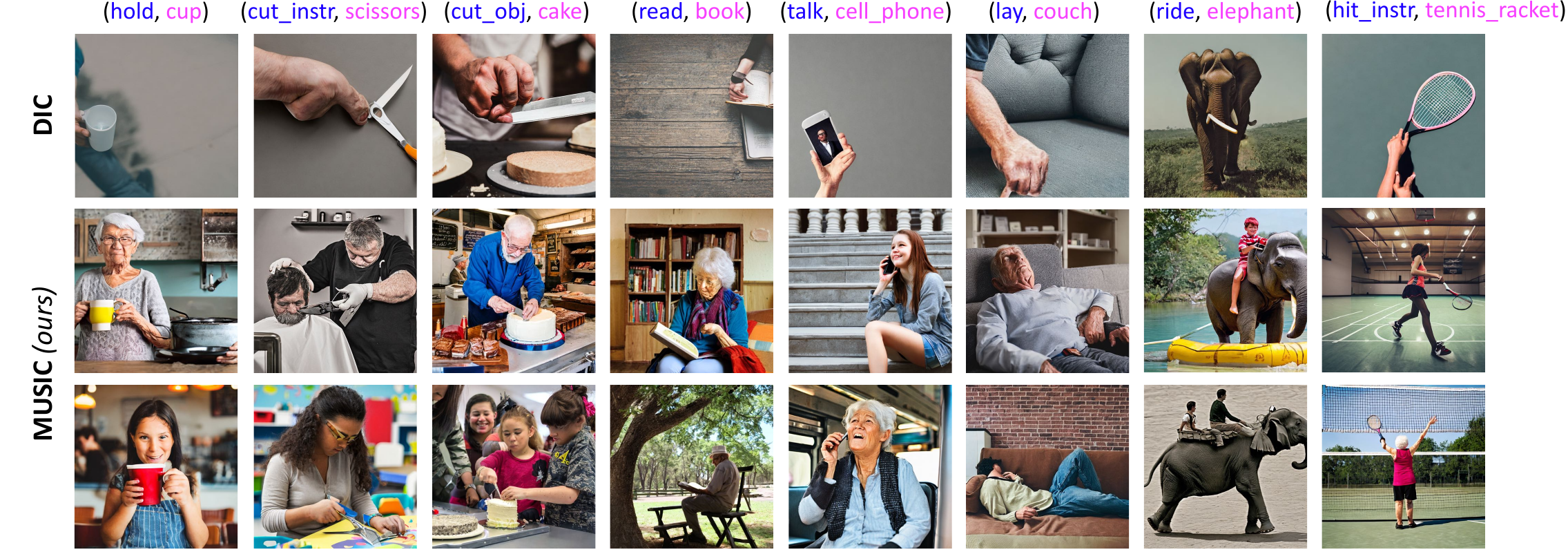}
\caption{
Comparison of virtual image samples generated by DIC (first row) and MUSIC (last two rows).
The interaction-object pair categories used to generate virtual images are marked at the top of images, 
where the interaction and object classes are in \textbf{BLUE} and \textbf{PINK}, respectively.
%
}
\label{fig:sd_vis}
\end{figure*}

%% file: fig/pseudo_labels.tex
\begin{figure*}[t]
\centering
\includegraphics[width=1.0\textwidth]{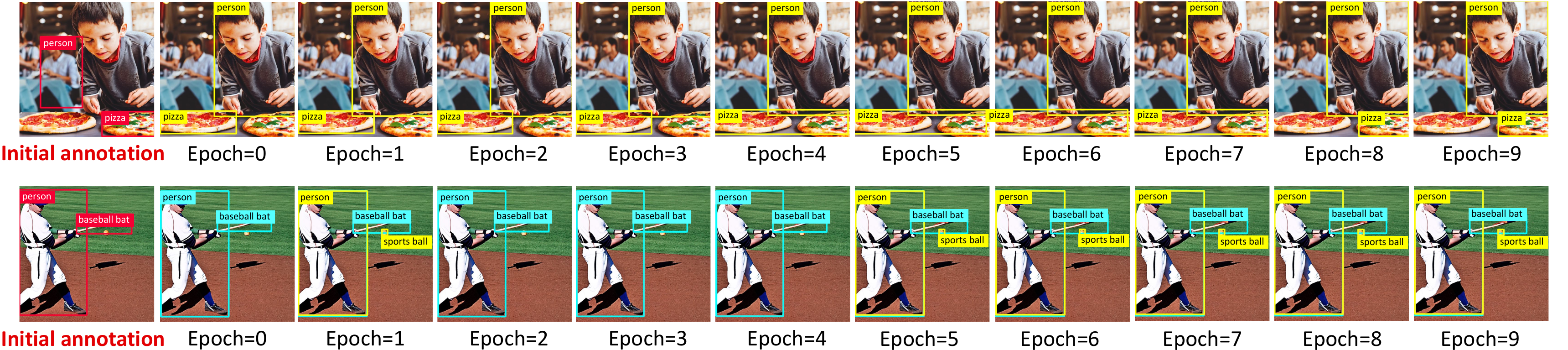}
\caption{
Visualization for the change process of the pseudo-labels.
We pick up some cases where the initial annotation created during the generation stage is inaccurate and show them on the leftmost with \textbf{RED}.
The pseudo-labels during training epochs are on the right, where the interactive pairs belonging to the same group are drawn with the same color.
}
\label{fig:pseudo_labels}
\end{figure*}

%% file: sec5_conclusion.tex
\section{Conclusion}
In this paper, we track the problem of long-tail the dilemma in HOI detection.
We propose a novel and general framework termed \textit{\textbf{V}irtual \textbf{I}mage \textbf{L}eaning} (VIL) to enhance existing HOI detectors.
In particular,
to generate a large-scale high-quality virtual dataset,
we design a \textit{\textbf{Mu}ltiple \textbf{S}teps \textbf{I}mage \textbf{C}reation} (MUSIC) approach.
Given an interaction-object pair category, 
MUSIC expands it to a plain sentence, polishes it with specific descriptions, and evaluates the generated image by multiple filtering steps.
In the training stage,
\textit{\textbf{A}daptive \textbf{M}atching-and-\textbf{F}iltering} (AMF) module is adopted to denoise and supplement the initial annotations of virtual images.
And the obtained pseudo-labels, along with the groud-truth, supervise the model learning knowledge from virtual and real images.
By combining five representative methods,
we evaluate the effectiveness and generalization of our VIL on two public datasets.
The results demonstrate all methods make significant progress with our method, and new state-of-the-art performances emerge.

%% file: sec_acknowledgement.tex
\section*{Acknowledgement}
This work was supported by National Key R\&D Program of China (No.2022ZD0118202), the National Science Fund for Distinguished Young Scholars (No.62025603), the National Natural Science Foundation of China (No. U21B2037, No. U22B2051, No. 62176222, No. 62176223, No. 62176226, No. 62072386, No. 62072387, No. 62072389, No. 62002305 and No. 62272401), and the Natural Science Foundation of Fujian Province of China (No.2021J01002,  No.2022J06001).